\documentclass[journal]{IEEEtran}
\usepackage{hyperref}
\usepackage{float}
\usepackage{verbatim} 
\usepackage{amssymb}
\usepackage{threeparttable}
\usepackage{multirow}
\usepackage{amsmath}
\usepackage{graphicx}
\usepackage{subfig}
\usepackage{epsfig} 
\usepackage{epstopdf}
\usepackage{diagbox}
\usepackage{booktabs} 
\restylefloat{figure}
\usepackage{color}
\usepackage{makecell}
\usepackage{bbding}
\usepackage{caption}
\begin{document}
%
\title{LGU-SLAM: Learnable Gaussian Uncertainty Matching with Deformable Correlation Sampling for Deep Visual SLAM}
%
%
%
%
\author{Yucheng~Huang,
        Luping~Ji, \emph{Member}, IEEE, Hudong Liu, Mao Ye, \emph{Senior Member}, IEEE
\thanks{This work was supported by the National Natural Science Foundation of
China (NSFC) under Grant 62476049. (\emph{Corresponding author: Luping Ji}.)}
\thanks{The authors are with the School 
of Computer Science and Engineering, University of Electronic Science and Technology of China, Chengdu 611731, China (email: 202411081508@std.uestc.edu.cn; jiluping@uestc.edu.cn; 3316950640@qq.com; cvlab.uestc@gmail.com).}

}

\maketitle

\begin{abstract}
Deep visual Simultaneous Localization and Mapping (SLAM) techniques, e.g., DROID, have made significant advancements by leveraging deep visual odometry on dense flow fields. In general, they heavily rely on global visual similarity matching. However, the ambiguous similarity interference in uncertain regions could often lead to excessive noise in correspondences, ultimately misleading SLAM in geometric modeling. To address this issue, we propose a Learnable Gaussian Uncertainty (LGU)
matching. It mainly focuses on precise correspondence construction. In our scheme, a learnable 2D Gaussian uncertainty model is designed to associate matching-frame pairs. It could generate input-dependent Gaussian distributions for each correspondence map.
Additionally, a multi-scale deformable correlation sampling strategy is devised to adaptively fine-tune the sampling of each direction by a priori look-up ranges, enabling reliable correlation construction. Furthermore, a KAN-bias GRU component is adopted to improve a temporal iterative enhancement for accomplishing sophisticated spatio-temporal modeling with limited parameters. The extensive experiments on real-world and synthetic datasets are conducted to validate the effectiveness and superiority of our method. Source codes are available at \href{https://github.com/UESTC-nnLab/LGU-SLAM}{https://github.com/UESTC-nnLab/LGU-SLAM.}
\end{abstract}


%
\IEEEpeerreviewmaketitle

\section{Introduction}
%
%
%
%
\IEEEPARstart{V}{isual} simultaneous localization and mapping, $i.e.$, visual SLAM, is realized through some sensors such as cameras to achieve the long-term pose tracking of intelligent agents and mapping of surrounding scenes in a structure-from-motion manner. Visual SLAM, as an important supporting technology, could greatly promote the application developments of autonomous driving, autonomous navigation, and virtual reality \cite{cadena2016past,alvarez2023monocular}.

\begin{figure}[t!]
	\centering
	
	\subfloat{
			\includegraphics[width=0.49\textwidth,height=0.32\textwidth]{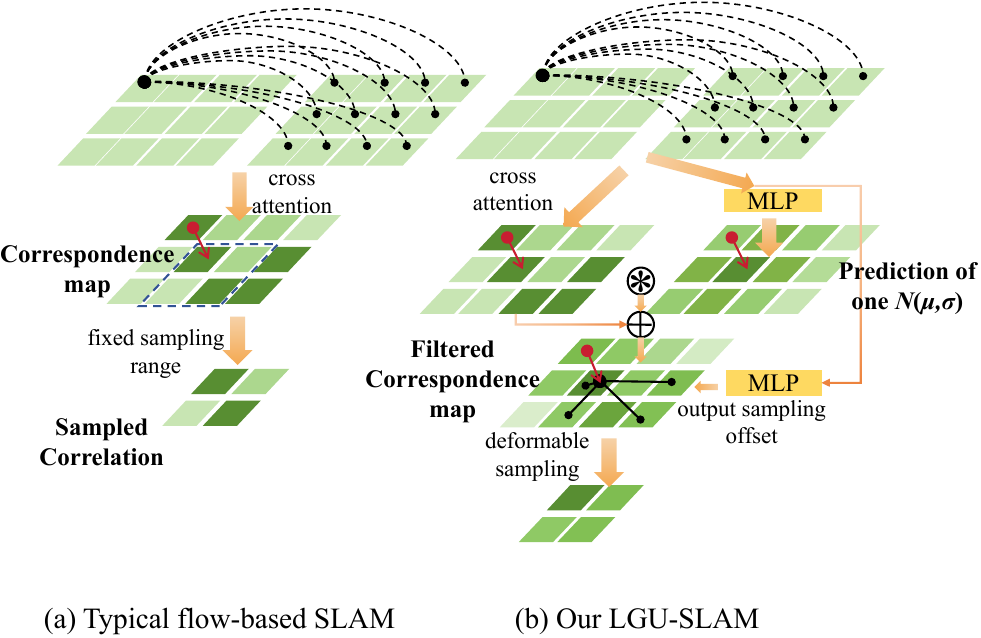}}
	\caption{Comparison of typical correspondence construction scheme and ours:
	(a) correlation volumes are directly established by correspondence maps through cross-attention, with a fixed range sampling; (b) our scheme adopts learnable Gaussian uncertainty to suppress the outliers in correlation volumes, with input-dependent deformable sampling to improve correlation context range.}
	\label{fig1} 
\end{figure}

Modern visual SLAM has adopted an optimization-based schemes\cite{klein2007parallel,newcombe2011dtam,forster2014svo,matsuki2018omnidirectional}, where camera pose and geometric mapping are jointly optimized through the bundle adjustment. Optimization-based SLAM could be roughly categorized into three classes according to different correspondence construction methods. The first class consists of feature matching methods such as the ORB-SLAM series\cite{mur2015orb,mur2017orb,campos2021orb}. Under the traditional approaches, sparse feature matching uses manual feature descriptors like ORB\cite{rublee2011orb} to construct correspondences. To improve semantic representation ability, relevant work\cite{sarlin2020superglue,teed2024deep,lipson2024deep} adopts deep learning for feature extraction and matching tasks. These indirect feature-matching methods have fast online tracking capabilities, while failure may occur in textureless areas. Simultaneously, due to the sparse feature matching, it cannot build dense maps. 

The second class consists of direct methods via photometric \cite{newcombe2011dtam,engel2014lsd,luo2018real}. These schemes could be executed viably in smooth areas, while the tremendous resolution leads to a catastrophic computational load. The third class is flow-based methods. They mine the correspondence between two frames by predicting the optical flow field while avoiding feature matching. DEMON\cite{ummenhofer2017demon} and DeepTAM\cite{zhou2018deeptam} use direct regression to predict optical flow and pose variations, resulting in unsatisfactory localization accuracy due to the lack of prior constraints. As a milestone in flow-based method, DROID-SLAM\cite{teed2021droid} utilizes visual correlation volumes to construct a dense optical flow field, and adopt deep dense bundle adjustment(DBA)\cite{tang2018ba} to optimize the depth and pose estimation iteratively. GO-SLAM\cite{zhang2023go} also uses correlation volumes to calculate the dense optical flow field for guiding pose optimization in the DBA layer and achieved high-quality dense mapping with NERF\cite{mildenhall2021nerf}.

Specially, typical flow-based methods, $e.g.$, DROID\cite{teed2021droid}, are built on the RAFT\cite{teed2020raft} optical flow algorithm. They construct a correlation volume by calculating the global visual similarity of input frames pairs and perform multi-scale sampling on the correlation volume using a priori specified range, as shown in Fig. \ref{fig1}(a). It is evident to identify potential flaws underlying the design. Firstly, in the process of global visual similarity calculation, each feature element in one frame attends to all elements in another frame equally, which does not accord with the prior perspective that the likelihood of each element in one previous frame appearing at various positions in the latter one should adhere to a 2D Gaussian distribution. 
Similarly, BRAFT \cite{jia2021braft} leverages this prior by assuming a fixed Gaussian distribution for each correspondence map, allowing for the manual filtering of distant but highly similar outliers. However, this assumption inevitably has a gap with real-world situations. Meanwhile, BRAFT cannot describe the uncertainty in ambiguous regions. Secondly, in these flow-based methods, the direction and step size of sampling at each scale are fixed. When the variance of the visual similarity distribution within this fixed sampling range is excessively large, it could lead to significant noise in sampling results. 

To address these issues, we propose a new LGU-SLAM scheme with reliable correspondence construction, as shown in Fig. \ref{fig1}(b). Inspired by 3DGS\cite{kerbl20233d}, in this scheme, we devise a learnable Gaussian uncertainty method for robust visual similarity matching. Our design motivation could be explained as follows. Firstly, in the calculation of correlation volumes, each element in the previous frame has a correspondence map, which is obtained through cross-attention with the latter one. We design learnable 2D Gaussian uncertain masks to weight each correspondence map, where the expectation and variance of each 2D Gaussian are input-dependent and predicted using multi-layer perceptron(MLP). We enable the center of the 2D Gaussian to move toward the reliable visual similarity region in the correspondence map conditioned on our devised uncertain self-supervised loss. To alleviate the negative impact of ambiguous regions in the correlation volumes, end-to-end Gaussian parameter adjustment is used to make ambiguous regions have a larger variance, thereby providing greater uncertainty and lower attention weights. To achieve a learnable 2D gaussian uncertain mask, we have designed the entire computing chain to be differentiable, so adjusting each 2D Gaussian in a posterior data-driven manner.

In addition, inspired by deformable convolution\cite{dai2017deformable}, we re-design a deformable multi-scale correlation sampling strategy to assist the model in looking-up correlation volume accurately. In our strategy, an offset decoder based on MLP is designed to predict the sampling offset of the top-level scale through the fusion of matched-frame pairs. A residual decoder is designed to predict the offset residual. It is added to the sampling offset of top-level scale and divided by the downsampling stride, to obtain the final sampling offset on the current scale. This allows for regularization using the prior information provided by the top-level offset to prevent the model from falling into sub-optimal optimization. Simultaneously, we utilize the uncertainty of the visual similarity distribution within the original sampling range to design a filtering mechanism for each predicted offset. Finally, we improved the key component GRU\cite{cho-etal-2014-properties} of the temporal iteration model in DROID by adding a KAN\cite{liu2024kan} based bias term to help it achieve more reliable complex temporal modeling under limited parameters. By integrating the proposed technical solutions to form the whole LGU-SLAM, robust real-time localization and mapping can be achieved.

The main contributions of our work could be summarized as follows:

(1) A learnable Gaussian uncertainty scheme is proposed to adjust the expectation and variance of each 2D Gaussian. It 
suppresses the interference from ambiguous regions by similarity uncertainty.

(2) A multi-scale deformable correlation sampling method is designed. It uses MLP to predict the sampling offset at each scale, facilitating the model to obtain reliable context when sampling correlation volumes.

(3) A KAN-bias GRU is adopted to achieve more reliable temporal modeling with limited parameters.

(4) The extensive experiments for our LGU-SLAM scheme on 4 benchmarks are conducted. They demonstrate the effectiveness and superiority of our scheme.

\section{Related work}
Our LGU-SLAM belongs to a kind of optimization scheme. There are three typical topics related to it, including feature matching, photometric-based, and flow-based method introductions.


\subsection{Feature matching method}
The indirect feature matching method needs to construct the feature descriptor of key points in the matching frame pairs for completing the correspondence construction. VINO-Mono\cite{qin2018vins} and $\text{OV}^{2}$SLAM\cite{ferrera2021ov} use binary robust independent without extra descriptors computation. ORB-SLAM\cite{mur2015orb} uses ORB features to complete the tracking and mapping of the whole slam system. Work\cite{yang2020robust} based on ORB features improves robustness to dynamic scenes through motion consistency of key points.

With the development of deep learning, the feature description is optimized in the off-line training process through the differentiable implicit neural representation. For instance, superglue\cite{sarlin2020superglue},EPM-Net\cite{zhang2023epm} and SCFeat\cite{sun2023shared} use neural networks to extract and match the feature points of the images, which are highly referential to the SLAM system. It is frustrating that the indirect method relies on background features with obvious texture, which makes it difficult to deal with smooth regions. At the same time, due to the sparse sampling of features, it is unable to achieve dense geometric mapping.

\subsection{Photometric based method}
The direct photometric-based method is first applied in DTAM\cite{newcombe2011dtam}, in which the similarity between the global pixel intensities of the matched frame pairs is used to construct a photometric loss to complete the optimization of pose estimation and mapping. LSD-SLAM\cite{engel2014lsd} realizes Gaussian Newton minimization and iteratively reduces the weight of large residuals, thus making tracking more robust against occlusions and reflections. 

Additionally, SVO\cite{forster2014svo} algorithm changes the error calculation of per-pixel into per-patch to prevent the model from falling into sub-optimization. Recently, high-quality dense mapping SLAM\cite{zhu2022nice,johari2023eslam,yan2024gs,keetha2024splatam}, hoping to achieve simultaneous optimization of tracking and mapping by using photometric loss and geometric loss. Nevertheless, these methods are difficult to achieve robust pose tracking, which indirectly leads to unreliable geometric modeling.

\subsection{Flow based method}
Flow-based methods construct the correspondence by establishing the optical flow field between matching pairs, thus replacing feature matching in supervising pose estimation. DytanVO\cite{shen2023dytanvo}, DEMON\cite{ummenhofer2017demon}, EDI-SLAM\cite{gong2020accurate} and DeepTAM\cite{zhou2018deeptam} directly predicting the optical flow field and pose through MLP, leading to lower tracking accuracy without prior constraints. 

To address this issue, the recently emerged advanced SLAM\cite{ye2022deflowslam,zhang2023go}, represented by DROID\cite{teed2021droid} establishes the optical flow fields by sampling the correlation volumes, and the DBA layer is used to optimize the pose and depth, realizing the impressive performance of deep SLAM and achieving good generalization. However, these methods do not consider the continuity and locality of the feature elements changing over time and space, easily suffering from the interference of distant anomalies. Moreover, the smooth region will lead to multitudinous ambiguous similarity points. Moreover, these methods adopt a fixed range, resulting in a limited ability to obtain reliable context.

Consequently, we present LGU-SLAM. By designing learnable Gaussian uncertainty, the spatio-temporal continuity and locality are taken into account. 2D Gaussian uncertainty masks are learned for each correspondence map using a posteriori-driven paradigm to suppress the outliers in the uncertain region. Furthermore, a spatio-temporal context-driven multi-scale deformable sampling method is designed to facilitate the model to obtain reliable context when sampling correlation volume. Finally, we propose the KAN-bias GRU improvement model to complete the complex temporal modeling under limited parameters.

\section{METHOD}
In this section, we detail the LGU-SLAM scheme, as shown in Fig. \ref{fig1}. The overall framework of the scheme is divided into three parts. The first is the deep feature extraction, responsible for reducing resolution while improving representation ability. The second part is correspondence construction. It leverages learnable Gaussian uncertainty to achieve high-confidence element-wise matching of frame pairs. Moreover, it employs multi-scale deformable correlation sampling to adaptively adjust the context mining range. The last is a temporal iterative module, performing complex temporal modeling with limited parameters by KAN-bias GRU.

\begin{figure*}[t!]
	\centering
	\subfloat{
		\includegraphics[width=0.8\textwidth,height=0.39\textwidth]{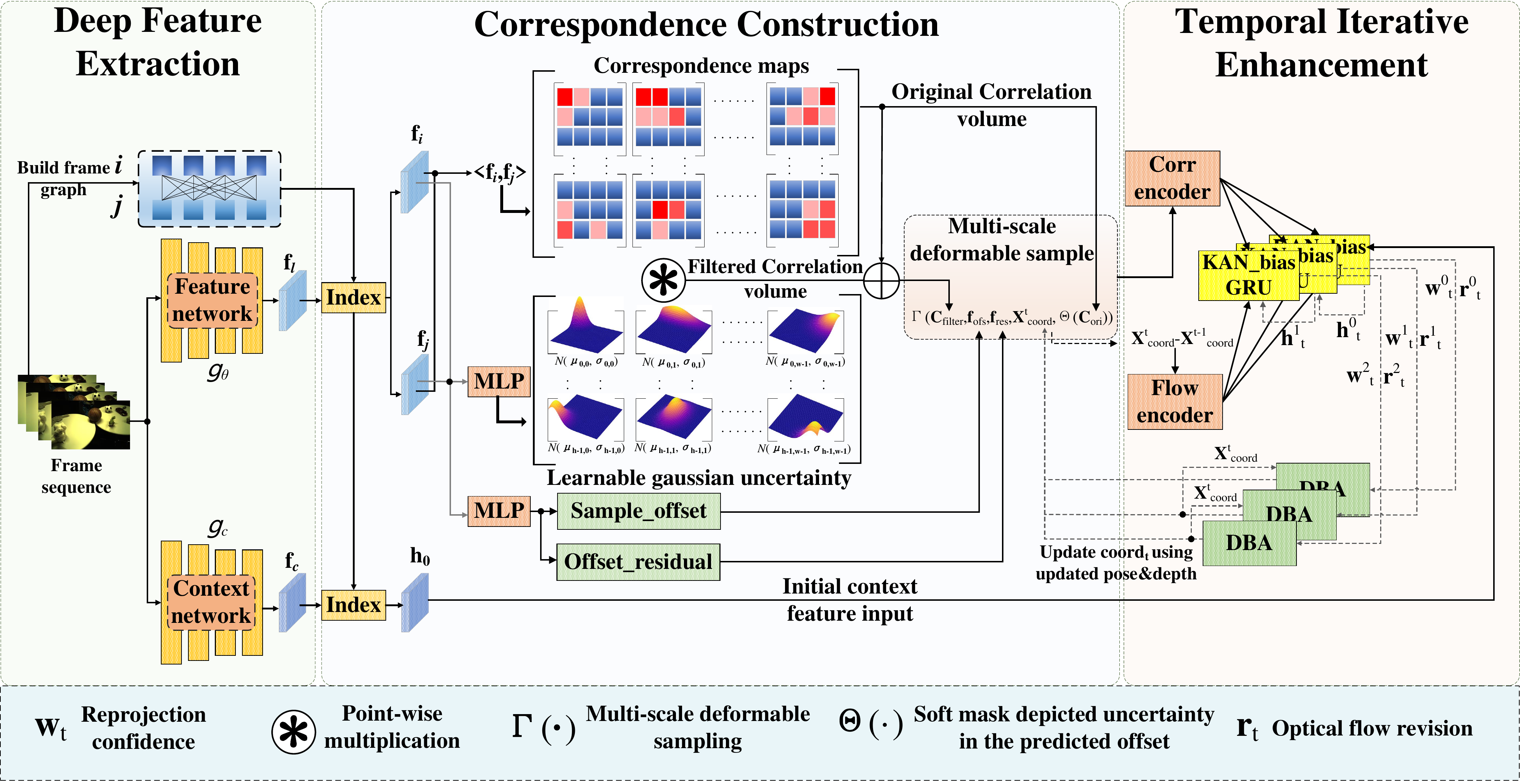}}
	\caption{Global Overview of our proposed LGU-SLAM. (1) Deep feature extraction. It utilizes video frames to input networks for deep semantic abstractions. (2) Correspondence construction. Firstly, It utilizes the bipartite graph to index the feature sequences $f_{l}$ for computing correlation volumes and devising an MLP based decoder outputs 2D Gaussian for all correspondence maps to generate Gaussian uncertainty masks that suppress outlier visual similarities. Secondly, It utilizes the proposed multi-scale deformable correlation sampling to enhance the contextual construction with the input-dependent sampling range. (3) Temporal iterative enhancement. It utilizes the designed KAN-bias GRU to perform temporal iterative enhancement, which is combined with dense bundle adjustment(DBA) for optimizing pose and depth information.}
	\label{fig2} 
\end{figure*}

\subsection{Deep Feature Extraction}
The input video frame sequence is processed using a convolutional neural network with shared parameters for feature extraction. Specifically, a feature network $g_{\theta}$, composed of six residual subnetworks, downsamples the original image with a stride of 8 to obtain the feature sequence $\mathbf{f}_{l}\in\mathbb{R}^{N\times H/8\times W/8 \times D}$, where $D$ is 128. Additionally, to provide historic context for the temporal iteration, a context network $g_{c}$ is established to process the video frames and obtain the context feature sequence $\mathbf{f}_{c}\in\mathbb{R}^{N\times H/8\times W/8 \times D}$. 

Simultaneously, the depth and pose of the input video frames are utilized to establish an optical flow field, measuring inter-frame distances based on the average optical flow magnitude, thus creating a co-visibility bipartite graph $\mathbf{G}_{b}\in\mathbb{R}^{2\times K}$($K$ is the number of matching pairs in $\mathbf{G}_{b}$) to index matching pairs sets $\{\left\langle \mathbf{f}^{k}_{i},\mathbf{f}^{k}_{j}\right\rangle\vert k=0, ..., K-1 \}$ in $\mathbf{f}_{l}$, where $i$ represents the previous 
moment and $j$ represents the subsequent moment. For $\mathbf{f}_{c}$, we only need to index the frameset of the previous moment $i$ to provide historical features for context construction.

\subsection{Correspondence Construction}\label{sec:lgu}
After obtaining the matching pairs set $\{\left\langle \mathbf{f}^{k}_{i},\mathbf{f}^{k}_{j}\right\rangle\vert k=0, ...,K-1 \}$ indexed by bipartite frame graph, 
we choose one matching pair $\left\langle \mathbf{f}^{k}_{i}\in\mathbb{R}^{C\times H\times W},\mathbf{f}^{k}_{j}\in\mathbb{R}^{C\times H\times W}\right\rangle$ from the matching set. Assuming that the pixel coordinates on the feature map $\mathbf{f}^{k}_{i}$ are $\mathbf{P}^{k}_{i}$, the camera pose of $\mathbf{f}^{k}_{i}$ is $\mathbf{T}^{k}_{i}\in SE(3)$, and its inverse depth is $\mathbf{d}^{k}_{i}\in\mathbb{R}^{H\times W}$. 
Then, the coordinate position of each pixel in $\mathbf{P}^{k}_{i}$, having been mapped into $\mathbf{f}^{k}_{j}$ using estimated pose and inverse depth, is computed by
\begin{equation}
\begin{aligned}
\mathbf{P}^{k}_{ij}=\Pi_{c}(\mathbf{T}^{k}_{ij}\circ {\Pi}^{-1}_{c}(\mathbf{P}^{k}_{i},\mathbf{d}^{k}_{i})) 
\\ \mathbf{T}^{k}_{ij}=\mathbf{T}^{k}_{j}\circ(\mathbf{T}^{k}_{i})^{-1}
\end{aligned}
\label{equation1}
\end{equation}
where $\Pi_{c}$ is the operation of projecting 3D points onto an image, ``$\circ$'' is a combination operation of two transformations, and ``-1'' indicates an inverse transformation.

When small camera motions occur between matching pairs, based on prior knowledge, we can provide a 2D Gaussian distribution $\mathbf{P}^{k}_{i}\sim{N(\mathbf{\mu}^{k}_{ij},\mathbf{\Sigma}^{k}_{ij})}$ to describe the probability of any given pixel coordinate in $\mathbf{P}^{k}_{i}$, appearing at various coordinate positions in $\mathbf{f}^{k}_{j}$. The probability density, $\mathcal{F}_{p}$ is estimated by 
\begin{equation}
\begin{aligned}
 \mathcal{F}_{p}(\mathbf{P}^{k}_{i})=\frac{\exp\left(-\frac{1}{2} (\mathbf{P}^{k}_{i} - \mathbf{\mu}^{k}_{ij})^T {\mathbf{\Sigma}^{k}_{ij}}^{-1} (\mathbf{P}^{k}_{i} - \mathbf{\mu}^{k}_{ij})\right)}{(2\pi)\sqrt{\text{det}(\mathbf{\Sigma}^{k}_{ij})}}
 \end{aligned}
\label{equation2}
\end{equation}
where "-1" still indicates an inverse transformation and $\text{det}(\cdot)$ is the determinant. If there is no camera motion between $\mathbf{f}^{k}_{i}$ and $\mathbf{f}^{k}_{j}$, then $\mathbf{P}^{k}_{i}$ equals $\mathbf{P}^{k}_{ij}$, and the expectation $\mathbf{\mu}^{k}_{ij}$ of this Gaussian distribution equals to pixel coordinates $\mathbf{P}^{k}_{i}$. 

Furthermore, if the area around one pixel in $\mathbf{P}^{k}_{i}$ has a high degree of smoothness, that is, the features in the area have an approximate visual similarity, then when the model relies on visual similarity to establish a correspondence, the 2D Gaussian distribution has a similar probability density and a large variance $\mathbf{\Sigma}^{k}_{ij}$ in this smooth area, which prompts the model to maintain a high degree of uncertainty in establishing correspondence. The above estimation of 2D Gaussian distribution parameters is based on posterior analysis, namely the camera movement (no motion) is known and there is a reliable feature extractor. Therefore, it is reasonable to construct the 2D Gaussian distribution through offline training for posterior driving.

\subsubsection{\textbf{Learnable Gaussian Uncertainty}}
Reviewing the 4D Correlation volume $\mathbf{C}\in\mathbb{R}^{
H\times W\times H\times W}$, which is composed of $H\times W$ correspondence maps. it is obvious that the visual similarity distribution in the correspondence maps conform to the 2D Gaussian distribution described above, because each correspondence map $\mathbf{C}_{m}\in\mathbb{R}^{H\times W}$ is an inner-product based global cross-attention of one pixel in $\mathbf{f}^{k}_{i}$ to all elements in $\mathbf{f}^{k}_{j}$. Hence, we can weight different elements in $\mathbf{C}_{m}$ based on probability, namely $\mathbf{C}_{m}=\mathbf{C}_{m}\ast \mathbf{P}_{m}$, where $\ast$ is point-wise multiplication, and $\mathbf{P}_{m}$ is the discrete probability obtained by sampling from 2D Gaussian $N(\mathbf{\mu}^{k}_{ij},\mathbf{\Sigma}^{k}_{ij})$. 

Since different correspondence maps $\mathbf{C}_{m}$ are obtained by globally paying attention to $\mathbf{f}^{k}_{j}$ through different pixels in $\mathbf{f}^{k}_{i}$, each $\mathbf{C}_{m}$ corresponds to a different 2D Gaussian. Taking inspiration from 3DGS\cite{kerbl20233d}, we design a learnable Gaussian uncertainty method and utilize neural networks to adaptively adjust all 2D Gaussian corresponding to $H\times W$ correspondence maps in the correlation volumes through an end-to-end data-driven approach. Ultimately, we achieve probabilistic weighting of regions within the matching range.

Given $H\times W$ correspondence maps $\mathbf{C}_{m}$, it is necessary to predict an expectation residual tensor $\mathbf{E}_{r}\in\mathbb{R}^{H\times W\times 2}$ and a covariance tensor $\mathbf{E}_{c}\in\mathbb{R}^{H\times W\times 4}$ separately. Due to the fact that each $\mathbf{C}_{m}$ is calculated from a matching pair, $\mathbf{f}^{k}_{i}$ and $\mathbf{f}^{k}_{j}$ achieve channel-wise concatenation $\text{Concat}_{cw}$ and input into a Gaussian parameter encoder $\mathcal{F}_{gs}$ constructed by a fully connected layer(FCN) to achieve multi-frame fusion. Then, the $\mathbf{E}_{r}$ and $\mathbf{E}_{c}$ are computed by 
\begin{equation}
\left\{
\begin{aligned}
\mathbf{E}_{r}=\mathcal{F}_{r}\bigg(\text{GeLU}\Big(\mathcal{F}_{gs}\big(\text{Concat}_{cw}(\mathbf{f}^{k}_{i},\mathbf{f}^{k}_{j})\big)\Big)\bigg)\\
 \mathbf{E}_{c}=\mathcal{F}_{c}\bigg(\text{GeLU}\Big(\mathcal{F}_{gs}\big(\text{Concat}_{cw}(\mathbf{f}^{k}_{i},\mathbf{f}^{k}_{j})\big)\Big)\bigg)
\end{aligned}
\right.
\label{equation3}
\end{equation}
where $\mathcal{F}_{r}$ is an expectation residual decoder and $\mathcal{F}_{c}$ is a covariance decoder, they both are built on the FCN. \text{GeLU} is an activation function.

To achieve a good trade-off between computational efficiency and performance, we define covariance as a diagonal matrix, hence, the final predicted covariance is $\mathbf{E}_{c}\in\mathbb{R}^{H\times W\times 2}$. In this way, during probability density calculations, matrix multiplication can be equivalent to point-wise multiplication, and the determinant can be obtained directly by multiplying diagonal elements, simultaneously, only two free variables remain in the covariance tensor to be predicted. 

After obtaining $\mathbf{E}_{c}$ through the $\mathcal{F}_{c}$, we perform correspondence normalization to ensure that the variance parameters of $H\times W$ 2D Gaussians are normalized under the same distribution, thereby improving the stability of model optimization. We leverage the sigmoid activation function to ensure the positive definiteness of the covariance matrix. So the final $\mathbf{E}_{c}$ is formulated as
\begin{equation}
\begin{aligned}
\mathbf{E}_{c}=\alpha\ast \Big(\text{Sigmod}\big(\text{Norm}_{corr}(\mathbf{E}_{c})\big)\Big)+\beta\\
\small
\text{Norm}_{corr}=\frac{\mathbf{E}_{c}-\text{mean}(\mathbf{E}_{c},\mathbf{NormIndex})}{\sqrt{\text{var}(\mathbf{E}_{c},\mathbf{NormIndex})+eps}}\\
\end{aligned}
\label{equation4}
\end{equation}
where $\ast$ is the point-wise multiplication, "mean" is the calculation of mean, "var" is the calculation of variance, and $\mathbf{NormIndex}$ refers to the dimension index in $\mathbf{E}_{c}$ that needs to participate in normalization calculation. In actual experiments, we manually set hyperparameters $\alpha=5$ and $\beta=0.05$ to scale the covariance parameters.

The final expectation tensor is calculated according to prior analysis, namely, we can obtain the Gaussian distribution center coordinates of one correspondence map $\mathbf{C}^{x,y}_{m}$ through indexing correlation volumes $\mathbf{C}$, and the coordinates could be close to $\mathbf{P}^{k}_{i}(x,y)_{idx}$($x\in[0,H],y\in[0,W]$). Therefore, we can obtain a prior basic expectation $\mathbf{E}_{b}$, which is the coordinates of the $\mathbf{f}^{k}_{i}$ feature grids. After adding $\mathbf{E}_{b}$ to the expectation residual $\mathbf{E}_{r}$, a 2D Gaussian expected tensor $\mathbf{E}_{\mu}=\mathbf{E}_{b}+\mathbf{E}_{r}$ is obtained. Ultimately, the probability density for each 2D Gaussian is obtained by
\begin{equation}
\begin{aligned}
\mathcal{F}_{p}(\mathbf{P}^{k}_{i})=\frac{\exp\big(-\frac{1}{2}\ast (\mathbf{P}^{k}_{i} - \mathbf{E}_{\mu})^{2}\ast {\mathbf{E}_{c}}^{-1}\big)}{2\pi\ast \big(\mathbf{E}_{c}(0)_{idx}\ast \mathbf{E}_{c}(1)_{idx}\big)^{1/2}}
\end{aligned}
\label{equation5}
\end{equation}
where "-1" indicates an inverse transformation, $(\cdot)_{idx}$ is the indexing operation, and $\ast$ is the point-wise multiplication. We sample the continuous probability density on the feature grids and multiply it by the hyperparameter($s=3$) as the final Gaussian mask $\mathbf{M}_{gs}$, which is applied to the original correlation volume $\mathbf{C}$, giving larger weight to deterministic regions within the matching range and less weight to uncertain regions($\mathbf{C}+=\mathbf{M}_{gs}\ast \mathbf{C}$). The entire calculation process is differentiable and can be easily integrated into the network for end-to-end learning.

In actual testing, we observed that the generated Gaussian mask is a sparse matrix with effective weights only within a very small range. To address this, we define a new Gaussian mask within a radius $r_{1}=(H+W)/(2*8)$ around the 2D Gaussian center to eliminate a substantial number of unnecessary zero elements. Consequently, the computational load of the Gaussian mask weights during forward propagation, as well as the gradient calculations for each element in the expectation tensor $\mathbf{E}_{\mu}$ and variance $\mathbf{E}_{c}$ tensor during backpropagation, are both reduced to ${1}/{r_{1}}$ of their original amounts. At the same time, we used PyTorch's CUDA extension to implement the entire learnable Gaussian uncertainty operator, which maintains the same frames per second(FPS) as DROID during the tracking and mapping stages.

\subsubsection{\textbf{Multi-Scale Deformable Correlation Sampling}}\label{sec:mds}
After filtering the correlation volume utilizing a learnable Gaussian uncertain mask, the next step is to use the average pooling with kernel sizes$\{1,2,4,8\}$ to downsample the correlation volume at four levels$\{\mathbf{C}_{1/s}\in\mathbb{R}^{H\times W\times H/s\times W/s} \vert s=1,2,4,8\}$. By sampling multi-scale correlation volume, the final correlation tensor $\mathbf{L}_{r}:\mathbb{R}^{H\times W\times H\times W}\times\mathbb{R}^{H\times W\times 2}\rightarrow \mathbb{R}^{H\times W\times (r+1)^{2}}$ is formed to guide the temporal iteration of subsequent update operators. 

The original multi-scale sampling in DROID, looks up the multi-scale volumes according to the coordinates $\mathbf{P}^{k}_{ij}$, that is, samples all elements within L1 distance with radius $r$ at each coordinate point in $\mathbf{P}^{k}_{ij}\in \mathbb{R}^{H\times W\times 2}$. 
DROID adopts a look-up approach, expecting to use visual similarity to find the potential coordinate position of any pixel of $\mathbf{f}^{k}_{i}$ in $\mathbf{f}^{k}_{j}$, in order to construct correspondence between the frame pair. The effectiveness of Look-up depends on whether the visual similarity within the sampling range constitutes a reliable context, but the predetermined sampling range in DROID limits the model's exploration of more reliable contextual information during offline training, especially in areas with obvious texture. 

It is natural to think of the sampling strategy of deformable convolution\cite{dai2017deformable}, but there are still differences in the looking-up between deformable convolution and correlation volume. Firstly, the original deformable convolution generates offset depending only on single-frame data. 

Although spatio-temporal deformable convolution\cite{deng2020spatio} is applied to temporal data, its generated offset is used for sampling in the three-dimensional feature space. We hope that the offset generated by temporal features can be sampled on the two-dimensional correlation map. Simultaneously, there is multi-scale sampling in the original process, which requires expanding the generated offset to the multi-scale level. Hence, we propose a multi-scale deformable sampling strategy for finding a solution.
\begin{figure}[t!]
	\centering
	\subfloat{
		\includegraphics[width=0.33\textwidth,height=0.39\textwidth]{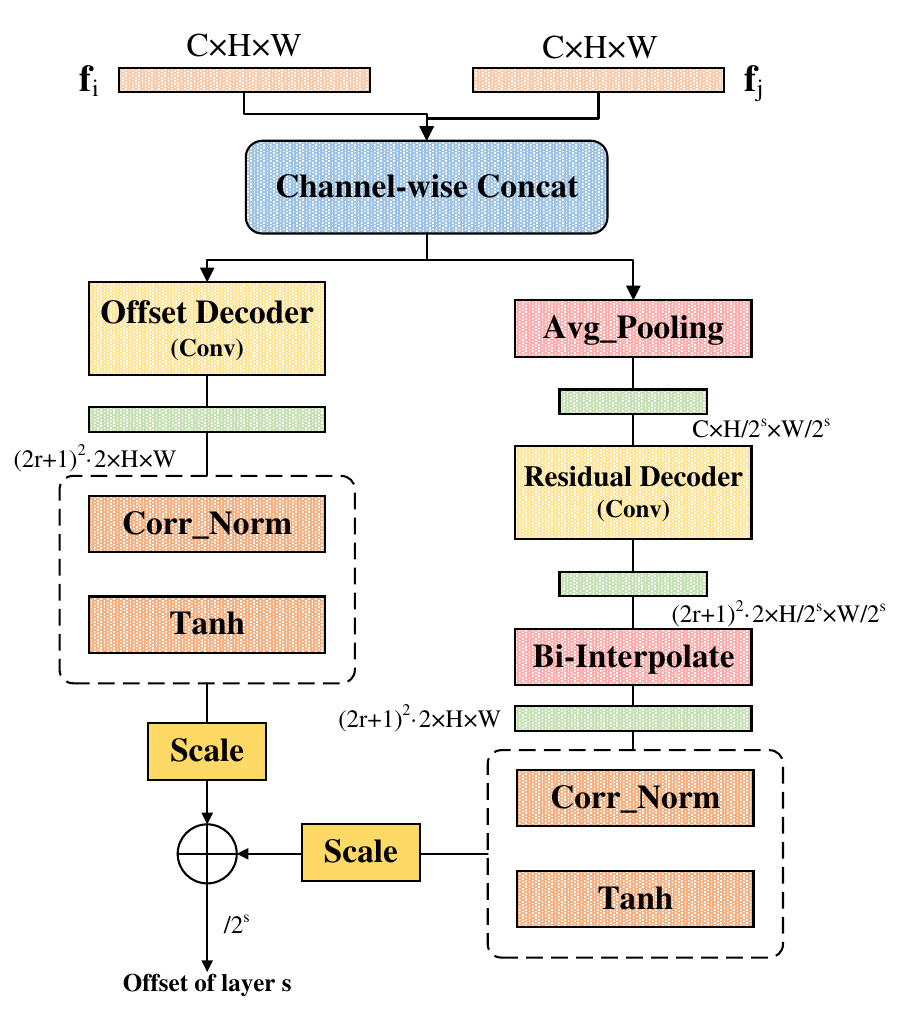}}
	\caption{Generation of multi-scale offset (layer s).}
	\label{fig3} 
\end{figure}

To construct multi-scale deformable sampling, temporal fusion is achieved firstly by channel-wise concatenation of $\mathbf{f}^{k}_{i}$ and $\mathbf{f}^{k}_{j}$, to output a two-dimensional sampling offset $\Delta_{ofs}\in\mathbb{R}^{(2r+1)^2\cdot2\times H\times W}$ using 2D convolution. As shown in Fig. \ref{fig3}, an offset decoder $\mathcal{F}_{ofs}$ for the top-level scale is implemented using convolutional layers. The original multi-scale sampling, starting from a priori, obtains the corresponding sampling coordinates for each scale by dividing $\mathbf{P}^{k}_{ij}\in \mathbb{R}^{H\times W\times 2}$ by the stride of the downsampling. Although bilinear interpolation can be used to approximate the sampling results of floating-point coordinates, it is still difficult to effectively align with the original scale. Therefore, when constructing multi-scale deformable sampling, we adopted a learnable residual structure for offset construction during downsampling. 

The prior basic offset of each lower scale is obtained through directly dividing the top-level offset $\Delta^{top}_{ofs}$ by the stride of the downsampling step. Then, a residual decoder $\mathcal{F}_{res}$ based on convolutional layers predicts the residual and adds it to the base offset through posterior driving. We can obtain the final offset $\Delta^{s}_{ofs}$ of the current layer(s) by
\begin{equation}
\small
\begin{aligned}
\Delta^{top}_{ofs}=\psi\Big(\mathcal{F}_{ofs}\big(\text{Concat}_{cw}(\mathbf{f}^{k}_{i},\mathbf{f}^{k}_{j})\big)\Big)\ast \tau_{s}\\
\Delta^{res}_{ofs}=\psi\bigg(\mathcal{I}_{b}\bigg(\mathcal{F}_{res}\Big(\text{AvgPool}\big(\text{Concat}_{cw}(\mathbf{f}^{k}_{i},\mathbf{f}^{k}_{j})\big)\Big)\bigg)\bigg)\ast \tau_{s}\\
\Delta^{s}_{ofs}=(\Delta^{top}_{ofs}+\Delta^{res}_{ofs})/2^{s}
\end{aligned}
\label{equation6}
\end{equation}
where $\psi$ is a combination of correspondence normalization $\text{Norm}_{corr}$ and \text{Tanh} activation function, further $\mathcal{I}_{b}$ is bilinear interpolation. $\text{Norm}_{corr}$ is consistent with section\ref{sec:lgu}. Finally, the hyperparameter $(\tau_{s}=4)$ is manually set to control the offset scale. Due to the significant impact of offset on the final sampling range and the lack of specific supervision signals for guidance, although end-to-end optimization can be achieved through gradient feedback, if not constrained by $\tau_{s}$, it could be subject to erroneous contextual interference in the early stages of training, falling into local minimum regions and leading to suboptimal performance.

\begin{figure}[t!]
	\centering
	\subfloat{
		\includegraphics[width=0.35\textwidth,height=0.35\textwidth]{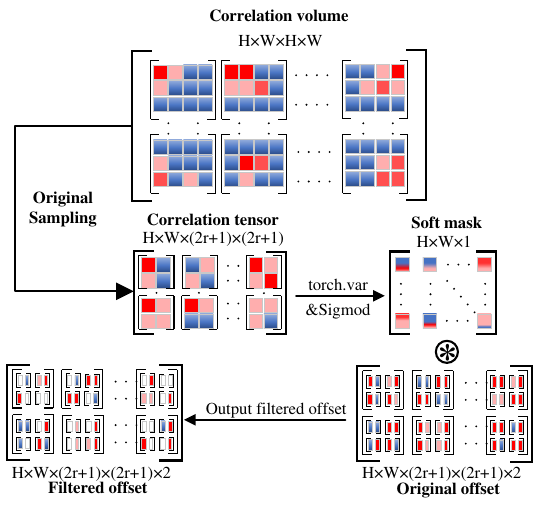}}
	\caption{Uncertainty based filtering in predicted offsets. To suppress redundant offsets, the final filtered offset is obtained by point-wise multiplication of the soft mask and the predicted offset tensor.}
	\label{fig4} 
\end{figure}

Besides, in areas with obvious texture, there is a richer context and a large variance of the visual similarity. On the contrary, in smooth regions, the variance of visual similarity distribution is small and it is difficult to form an effective background context. If we sample using the generated offset in this area, reliable information could not be mined. 

Since in smooth regions, we can already determine that the surrounding visual features are approximate, resulting in a deterministic construction of the context, hence fixed range sampling is sufficient. So we design a filtering mechanism for each predicted offset tensor, as depicted in Fig. \ref{fig4}, using the uncertainty of the similarity distribution of the original sampling range as the filtering threshold, and constructing a differentiable soft mask to suppress the interference of redundant offsets on model performance. The formula represents the generation of uncertainty-based filtering is
\begin{equation}
\begin{aligned}
\Theta(C)=\text{var}\big(\Psi(\mathbf{C}),\mathbf{VarIndex}\big)\\
\Delta_{ofs}=\text{Sigmod}\big(\Theta(C)\big)*\Delta_{ofs}
\end{aligned}
\label{equation7}
\end{equation}
where $\Psi$ is the original sampling for $\mathbf{C}$ to obtain the correlation tensor $\mathbf{L}^{ori}_{r}\in\mathbb{R}^{H\times W\times (r+1)^{2}}$, further we calculate the variance of $H\times W$ values with a sampling radius of $r$ in $\mathbf{L}^{ori}_{r}$, $\mathbf{VarIndex}$ refers to the dimension index in $\mathbf{L}^{ori}_{r}$ that needs to participate in variance calculation, finally generates a soft mask through \text{sigmod} and multiplies it with $\Delta_{ofs}$. Due to the use of a soft paradigm, this part can be optimized with the entire model. 

After obtaining the filtered $\mathbf{C}$ and $\Delta_{ofs}$, we can proceed with the final deformable sampling. Among them, the sampling operation is consistent for each scale. Given that $(px, py)$ is a coordinate in $\mathbf{P}^{k}_{ij}$, the original sampling range is $\{px+i, py+j\vert-r<i<r, -r<j<r\}$. For one $(px, py)$, We can extract $(2r+1)^{2}$ coordinate offsets from the predicted offset tensor$\Delta_{ofs}$, by adding to the original sampling range, the final sampling range can be obtained $\{px+i+\Delta_{ofs}(ix), py+j+\Delta_{ofs}(jy)\vert-r<i<r, -r<j<r\}$, where $(ix)$ represents indexing the i-th element of x-dim in $\Delta_{ofs}$, and $(iy)$ represents similarly. By combining the final sampling range and using bilinear interpolation, sampling on the correlation volume can be achieved. Similarly, we use PyTorch's CUDA extension to assist in implementing multi-scale deformable sampling operators, improving the overall computational speed of the model. 

As the back-end optimization contains massive key frames, the original correlation sampling has high GPU-memory consumption. In this regard, the correlation sampling method in RAFT is used for reference. The original multi-scale deformable correlation sampling consists of three stages, first calculating the correlation volumes, then pooling the correlation volumes in multi-scale, and finally executing deformable sampling. Unfortunately, the calculation of correlation volumes will cause the computational complexity of $C\ast (H\ast W)^{2}$. Assuming channel $C$ is constant, it will bring quadratic complexity with the increase of $H$ and $W$. After deformable sampling calculation, the final computational complexity is $O_{1}((H\ast W)^{2})$. 

Instead, the correlation can be directly calculated by deformable sampling between the two feature maps. In this way, the whole calculation overhead is reduced to $C\ast H\ast W\ast(2r+1)^{2}$, and the linear complexity $O_{2}(H\ast W)$ is realized. Since the computational complexity of pooling is linear for $H*W$, if we take pooling calculations into account, $O_{1}$ is still much greater than $O_{2}$.

\subsection{Temporal Iterative Enhancement}\label{sec:kan}
After constructing a correspondence using visual similarity, we can obtain a correlation tensor $\mathbf{L}_{r}$, which will be input into the update operator along with the initial hidden state $\mathbf{h}_{0}$ output by the context network for temporal iteration, completing the update of the hidden state $\mathbf{h}_{t}$ and optical flow field. The core component of the update operator is \text{GRU}. We adopt the KAN with learning ability in the activation unit to generate bias for the update gate, reset gate, and output gate in \text{GRU}. Through nonlinear learnability, we effectively map low-dimensional features to a higher-dimensional space, thereby improving the semantic discrimination ability of the model and achieving the goal of effectively filtering and preserving historical information.

Specifically, KAN is used to predict biases for update gates, reset gates, and output gates in GRU. By inputting hidden state $\mathbf{h}_{t-1}$, bias terms corresponding to the gate units are obtained. These bias terms are then subjected to point-wise addition with the output results of the three gate units to constrain the information flow of each gate unit, so the core component KAN-bias GRU can be computed by
\begin{equation}
\small
\begin{aligned}
\mathbf{z}_{t}=\text{Sigmod}\big(\text{Conv}_{3\times3}([\mathbf{h}_{t-1},\mathbf{x}_{t}],\mathbf{W}_{z})+\mathbf{b}^{z}_{kan}\big)\\
\mathbf{r}_{t}=\text{Sigmod}\big(\text{Conv}_{3\times3}([\mathbf{h}_{t-1},\mathbf{x}_{t}],\mathbf{W}_{r})+\mathbf{b}^{r}_{kan}\big)\\
\mathbf{o}_{t}=\text{Tanh}\big(\text{Conv}_{3\times3}([\mathbf{r}_{t}\ast\mathbf{h}_{t-1},\mathbf{x}_{t}],\mathbf{W}_{o})+\mathbf{b}^{o}_{kan}\big)\\
\mathbf{h}_{t}=(1-\mathbf{z}_{t})\ast\mathbf{h}_{t-1}+\mathbf{z}_{t}\ast\mathbf{o}_{t}\\
\mathbf{b}^{'}_{kan}=\text{KAN}^{'}\Big(\text{Sigmod}\big(\text{Conv}_{1\times1}(\mathbf{h}_{t-1})\big)\ast\mathbf{h}_{t-1}\Big)
\end{aligned}
\label{equation8}
\end{equation}
where $\mathbf{W}_{z}$, $\mathbf{W}_{r}$, $\mathbf{W}_{o}$ are the weights of convolution layers.
\subsection{Training and Loss Definition}\label{sec:sys}
\textbf{Training.} During the training phase, we set the batch size to 2, and each training sample subset consists of 4 consecutive video frames. Consistent with DROID, the average optical flow field between video frames is calculated to construct a distance matrix. Since not every frame pair necessarily satisfies co-vision, only sample pairs with distances less than the specified threshold (d=24) in the distance matrix are selected to form the final frame graph.

To guide the optimization of the entire network model, we constructed two supervised loss functions and two self-supervised loss functions. Firstly, given the groundtruth pose and predicted pose, the $\mathcal{L}_{pose}$ is measured by the differential between the two. Secondly, using the given groundtruth pose and depth map to construct a groundtruth optical flow field, which is used to calculate L1 loss with the predicted optical flow field to get the optical flow loss $\mathcal{L}_{flow}$. Thirdly, a residual loss $\mathcal{L}_{res}$ is constructed between the $\mathbf{P}^{k}_{ij}$ before and after DBA optimization during the temporal iteration process to fine-tune the update operator network. The above three loss functions are formulated as in DROID\cite{teed2021droid}. Finally, we propose the uncertain self-supervised loss $\mathcal{L}_{self}$ for the parameter updates in learnable Gaussian uncertainty.

Without the constraint of $\mathcal{L}_{self}$, the prediction of expectation residuals is highly likely to fall into sub-optimal states. This means that the prediction of the 2D Gaussian center positions, despite deviating from the optical flow field, may still guarantee the local minimization of the overall system loss, such a scenario poses a catastrophic threat to the generalization capability of the system. We draw on the uncertainty model proposed in \cite{kendall2017uncertainties} to construct a self-supervised loss for the expectation residual decoder and the variance decoder, and the loss can be computed by
\begin{equation}
\begin{aligned}
\mathcal{L}_{self}=\dfrac{1}{H*W*2*\text{det}(\mathbf{E}_{c})}\lvert\lvert \mathbf{E}_{\mu}-\mathbf{P}^{k}_{ij}\rvert\rvert^{2}\\
+\dfrac{1}{2}\text{log}\big(\text{det}(\mathbf{E}_{c})\big)
\end{aligned}
\label{equation9}
\end{equation}
where $\text{det}(\cdot)$ is the determinant of the $\mathbf{E}_{c}$, $\mathbf{P}^{k}_{ij}$ as a self-supervised signal characterizes the optical flow field between two feature maps. Our objective is for the predicted expectation residual to align the center of the 2D Gaussian corresponding to each correspondence map with the position of each pixel in the other frame. Consequently, we utilize the $\mathbf{P}^{k}_{ij}$ obtained after multiple iterations using a temporal model to self-supervise the parameter optimization of the expectation residual decoder $\mathbf{E}_{\mu}$. The determinant of the 2D Gaussian covariance $\mathbf{E}_{c}$ corresponding to each correspondence map is employed to measure the current correspondence uncertainty, thereby suppressing the gradient values in regions with high noise levels and preventing invalid updates to the parameters.

The total loss of LGU-SLAM is obtained by
\begin{equation}
\begin{aligned}
\mathcal{L}_{total}=\lambda_{1}\mathcal{L}_{pose}+\lambda_{2}\mathcal{L}_{flow}+\lambda_{3}\mathcal{L}_{res}+\lambda_{4}\mathcal{L}_{self}
\end{aligned}
\label{equation10}
\end{equation}

\textbf{Inference.} Consistent with DROID, the entire LGU-SLAM system is divided into two threads, front-end and back-end. Before the system runs, accumulate 12 frames for initialization to build the initial frame graph. After the tracking starts, the SLAM front-end continuously receives video input frames and constructs a frame graph composed of keyframes in the co-vision frame set. Each time, a frame is selected from the three nearest neighbors of the last added video frame based on the average optical flow magnitude and added to the frame graph. 

Throughout the tracking process, the pose and depth are initialized through a fixed linear transformation, and the network parameters of the update operator are fixed. Local BA is used to iteratively optimize the pose and depth on the frame graph. Once the number of keyframes exceeds the window size, the average optical flow size of each frame pair in the frame graph will be calculated, and the frame with the longest distance or the older one will be removed. The backend receives the overall historical keyframes and optimizes the entire frame graph through global BA iteration.

\begin{table*}[!ht]
     \caption{Comparison results on the TartanAir Monocular benchmark, ATE[m].}  
	\centering
	\renewcommand{\arraystretch}{1}
	\setlength{\tabcolsep}{6pt}
	\label{table1}
	\begin{threeparttable}
		\resizebox{\linewidth}{!}{
			\begin{tabular}{lcccccccc|cccccccc|cc}\Xhline{1.5pt}
				\multicolumn{1}{l}{\multirow{2}{*}{Methods}} &ME &ME &ME &ME&ME&ME&ME&ME&MH&MH&MH&MH&MH&MH&MH&MH&\multirow{2}{*}{Avg} \\
                \multirow{2}{*}{ } &000 &001 &002 &003 &004&005&006&007&000 &001 &002 &003 &004&005&006&007&\multirow{2}{*}{ }\\\hline
                ORB-SLAM3\cite{cadena2016past} &13.61 &16.86 &20.57 &16.00 &22.27&9.28&21.61&7.74&15.44&2.92 &13.51 &8.18 &2.59&21.91&11.70&25.88&14.38\\
                COLMAP\cite{schonberger2016structure} &15.20 &5.58 &10.86 &3.93 &2.62&14.78&7.00&18.47&12.26&13.45 &13.45&20.95 &24.97 &16.79&7.01&7.97&12.50\\
                DSO\cite{engel2017direct}&9.65 &3.84 &12.20 &8.17 &9.27&2.94&8.15&5.43&9.92&0.35 &7.96 &3.46 & \XSolidBrush &12.58&8.42&7.50&7.32\\
                DROID-SLAM\cite{teed2021droid}&0.27 &1.06 &0.36 &0.87 &2.14&0.13&1.13&\textbf{0.06}&\textbf{0.08}&\textbf{0.05}&0.04&0.02 &1.30 & 0.68 &0.30&\textbf{0.07} &0.53\\
                DPVO\cite{teed2024deep}&0.16 &\textbf{0.11}&\textbf{0.11} &0.66 &\textbf{0.31} &0.14&\textbf{0.64}&0.13&0.21&\textbf{0.05}&0.04&0.08 &\textbf{0.58} & \textbf{0.17} &\textbf{0.11} &0.35 &\textbf{0.24}\\
                DeFlowSLAM\cite{ye2022deflowslam}&- &-&- &- &- &-&-&-&0.63&0.06 &\textbf{0.02} &\textbf{0.01} &2.80 &0.20&0.31&0.45 &-\\ \hline
                LGU-SLAM(Ours)&\textbf{0.08} &0.13 &0.30 &\textbf{0.43} &0.70 &\textbf{0.07} &0.81 &0.09 &0.09&\textbf{0.05} &\textbf{0.02} &0.06 &1.03 &0.34 &0.26& 0.16 &0.29\\
    \Xhline{1.5pt}
			\end{tabular}
		}	
	\end{threeparttable}
\end{table*}

\begin{figure*}[htbp]
  \begin{minipage}{0.5\textwidth}
        \centering
	\subfloat{
		\includegraphics[width=0.8\textwidth,height=0.5\textwidth]{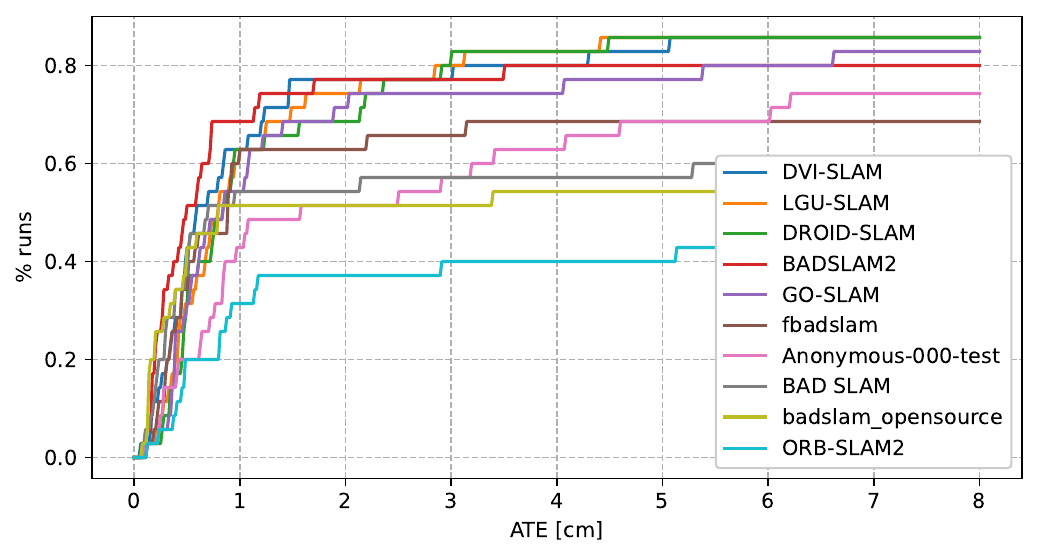}} 
  \end{minipage}
  \hfill 
  \begin{minipage}{0.50\textwidth}
  \centering
  \footnotesize
  \renewcommand{\arraystretch}{1.1}
	\setlength{\tabcolsep}{2.5pt}
    \begin{tabular}{l|c|cc}
      \Xhline{1.2pt}
      \multicolumn{1}{c|}{\multirow{2}{*}{Methods}} & \multicolumn{1}{c|}{\multirow{2}{*}{Input}} & \multicolumn{2}{c}{\multirow{1}{*}{AUC(ETH3D-test) $\uparrow$}} \\
      \multicolumn{1}{c|}{\multirow{2}{*}{}} & \multicolumn{1}{c|}{\multirow{2}{*}{}} & Max error=8cm& Max error=2cm \\\hline
      BundleFusion\cite{dai2017bundlefusion}&RGB-D&33.84&2.59\\
      ElasticFusion\cite{whelan2015elasticfusion}&RGB-D&34.02&2.69\\
      DVO-SLAM\cite{kerl2013dense}&RGB-D&71.83&5.83\\
      ORB-SLAM3\cite{cadena2016past}&RGB-D&73.33&5.82\\
      ORB-SLAM2\cite{mur2017orb}&RGB-D&104.28&17.84\\
      BAD-SLAM\cite{schops2019bad}&RGB-D&153.47&30.89\\
      GO-SLAM\cite{zhang2023go}&RGB-D&197.02&33.12\\
      DROID-SLAM\cite{teed2021droid}&RGB-D&207.79 &32.84\\
      DVI-SLAM\cite{peng2024dvi}&RGB-D&211.60 &37.97\\\hline
     Our LGU-SLAM &RGB-D&209.89&34.42\\
      \Xhline{1.2pt}
    \end{tabular}
  \end{minipage}
  
  \caption{Comparison results on the ETH3D-test RGB-D benchmark, ATE[cm]. Our method can achieve excellent generalization effect without fine-tuning on ETH3D like DVI-SLAM, and the test results can be found in \href{https://www.eth3d.net/slam_benchmark}{ETH3D-official.}}
  \label{fig6}
\end{figure*}

\section{Experiments}
To verify the effectiveness of LGU-SLAM, we conduct robustness validation on various types of datasets, including synthetic datasets such as TartanAir dataset, TartanAir Visual SLAM challenge\cite{wang2020tartanair}, and ETH3D-SLAM\cite{schops2019bad}, as well as real-world datasets such as EuRoC\cite{burri2016euroc} and TUM-RGB\cite{sturm2012benchmark}. According to the evaluation criteria of previous work, we select absolute trajectory error ATE and area under the curve(AUC) as the measurement standards for SLAM pose estimation and tracking capability. Meanwhile, we select SOTA methods from deep learning-based visual SLAM and traditional visual SLAM tasks for comparative testing to validate the effectiveness and competitiveness of LGU-SLAM. Finally, ablation analysis is used to verify the impact of the proposed method on overall system performance.

\subsection{implementation Details}
LGU-SLAM is trained using an NVIDIA RTX 3090 GPU, and the training dataset consists entirely of the TartanAir dataset. To facilitate model validation, we have divided the training and validation sets. The overall training requires 250K steps, with batch size set to 2 and a video frame length of 4 input to the network each time. The pose of the first two frames is fixed as groundtruth to remove gauge freedom. The input frame resolution is $384\times512$, and 8 temporal iterations are performed each time to accumulate errors and optimize the model. In the loss calculation process, the weight $\lambda_{1}$ of $\mathcal{L}_{pose}$ is 10, the weight $\lambda_{2}$ of $\mathcal{L}_{flow}$ is 0.05, the weight $\lambda_{3}$ of $\mathcal{L}_{res}$ is 0.01, and the weight $\lambda_{4}$ of $\mathcal{L}_{e}$ is 0.08.

 \begin{table*}[!ht]
        \caption{Comparison results on the EuRoC benchmark, ATE[m].}
	\centering
	\renewcommand{\arraystretch}{1}
	\setlength{\tabcolsep}{10pt}
	\label{table2}
	\begin{threeparttable}
		\resizebox{\linewidth}{!}{
			\begin{tabular}{cl|ccccc|ccc|ccc|c}\Xhline{1.5pt}
				\multicolumn{2}{c|}{\multirow{1}{*}{VO/SLAM}} &MH01 &MH02 &MH03 &MH04&MH05&V101&V102&V103&V201&V202&V203&Avg \\
                \hline
                \multicolumn{1}{c}{\multirow{7}{*}{Mono VO}} &DSO\cite{engel2017direct} &\textbf{0.046} &\textbf{0.046} &0.172&3.810&\textbf{0.110}&0.089&\textbf{0.107}&0.903&0.044&0.132&1.152&0.601\\ 
                \multicolumn{1}{c}{\multirow{7}{*}{ ATE(m)$\downarrow$}}&SVO\cite{forster2014svo}&0.100&0.120&0.410&0.430&0.300 &0.070 &0.210 &\XSolidBrush &0.110&0.110&1.080&-\\
                \multicolumn{1}{c}{\multirow{7}{*}{}}&DeepV2D\cite{Teed2020DeepV2D} &0.739 &1.144 &0.752 &1.492 &1.567 &0.981&0.801&1.570&0.290&2.202 &2.743&1.298\\
                \multicolumn{1}{c}{\multirow{7}{*}{}}&TartanVO\cite{wang2021tartanvo}&0.639 &0.325 &0.550 &1.153 &1.021&0.447&0.389&0.622&0.433&0.749 &1.152 &0.680\\
                \multicolumn{1}{c}{\multirow{7}{*}{}}&DROID-VO\cite{teed2021droid}&0.163 &0.121 &0.242 &0.399 &0.270&0.103&0.165&0.158&0.102&0.115&\textbf{0.204}&0.179\\
                \multicolumn{1}{c}{\multirow{7}{*}{}}&DVI-VO\cite{peng2024dvi}&0.120 &0.063&\textbf{0.076} &\textbf{0.161} &0.354 &\textbf{0.060}&0.207&\textbf{0.113}&0.087&\textbf{0.061}&0.268&0.143\\
                \multicolumn{1}{c}{\multirow{7}{*}{}}&LGU-VO(Ours)&0.107 &0.089&0.136 &0.232 &0.267 &0.083&\textbf{0.125}&0.128&\textbf{0.040}&0.067&0.264&\textbf{0.139}\\\hline
                
                
                \multicolumn{1}{c}{\multirow{5}{*}{Mono SLAM}} &DeepFactors\cite{czarnowski2020deepfactors} &1.587 &1.479 &3.139&5.331&4.002&1.520&0.679&0.900&0.876&0.905&1.021&2.040\\ 
                \multicolumn{1}{c}{\multirow{5}{*}{ ATE(m)$\downarrow$}}&ORB-SLAM\cite{mur2015orb}&0.071&0.067&0.071&0.082&0.060 &0.015 &0.020 &\XSolidBrush&0.021&0.018&\XSolidBrush&-\\
                \multicolumn{1}{c}{\multirow{5}{*}{}}&ORB-SLAM3\cite{campos2021orb} &0.016 &0.027 &0.028 &0.138 &0.072 &\textbf{0.033}&0.015&0.033&0.023&0.029 &\XSolidBrush&-\\
                \multicolumn{1}{c}{\multirow{5}{*}{}}&DROID-SLAM\cite{teed2021droid}&0.013 &0.014 &0.022 &0.043 &0.043&0.037&0.012&0.020&0.017&0.013 &0.014 &0.022\\
                \multicolumn{1}{c}{\multirow{5}{*}{}}&DPV-SLAM\cite{teed2021droid}&0.013 &0.016 &0.021 &0.041 &0.041&0.035&0.010&\textbf{0.015}&0.021&0.011 &0.023 &0.023\\
                \multicolumn{1}{c}{\multirow{7}{*}{}}&LGU-SLAM(Ours)&\textbf{0.010} &\textbf{0.012} &\textbf{0.016} &\textbf{0.033} &\textbf{0.039}&0.038&\textbf{0.011}&0.018&\textbf{0.014}&\textbf{0.009}&\textbf{0.011}&\textbf{0.018}\\
                
    \Xhline{1.5pt}
			\end{tabular}
		}	
	\end{threeparttable}
\end{table*}

 \begin{table*}[!ht]
        \caption{Comparison results on the TUM-RGBD benchmark, ATE[m].}
	\centering
	\tiny
	\renewcommand{\arraystretch}{1}
	\setlength{\tabcolsep}{10pt}
	\label{table3}
	\begin{threeparttable}
		\resizebox{\linewidth}{!}{
			\begin{tabular}{l|ccccccccc|c}\Xhline{0.8pt}
				 &360 &desk &desk2 &floor&plant&room&rpy&teddy&xyz&Avg \\
                \hline
                ORB-SLAM2\cite{mur2017orb}&\XSolidBrush &0.071 &\XSolidBrush &0.023&\XSolidBrush&\XSolidBrush&\XSolidBrush&\XSolidBrush&0.010&-\\ 
                ORB-SLAM3\cite{campos2021orb}&\XSolidBrush&0.017&0.210&\XSolidBrush&0.034 &\XSolidBrush &\XSolidBrush &\XSolidBrush &0.009&-\\\hline
                 DeepTAM\cite{zhou2018deeptam}&\textbf{0.111}&0.053&0.103&0.206&0.064 &0.239 &0.093 &0.144 &0.036&0.116\\
                  TartanVO\cite{wang2021tartanvo}&0.178&0.125&0.122&0.349&0.297 &0.333 &0.049 &0.339 &0.062&0.206\\
                   DeepV2D\cite{Teed2020DeepV2D}&0.243&0.166&0.379&1.653&0.203 &0.246 &0.105 &0.316 &0.064&0.375\\
                   DeepFactors\cite{czarnowski2020deepfactors}&0.159&0.170&0.253&0.169&0.305 &0.364 &0.043 &0.601 &0.035&0.233\\
                   DeFlowSLAM\cite{ye2022deflowslam}&0.159&0.016&0.030&0.169&0.048 &0.538 &0.021 &0.039 &0.009&0.114\\
                   GO-SLAM\cite{zhang2023go}&0.089&0.016&0.028&0.025&0.026 &0.052 &0.019 &0.048 &0.010&0.035\\
                   DPV-SLAM\cite{lipson2024deep}&0.112&0.018&0.029&0.057&0.021 &0.330 &0.030 &0.084 &0.010&0.076\\
                   DROID-SLAM\cite{teed2021droid}&\textbf{0.111}&0.018&0.042&0.021&0.016 &0.049 &0.026 &0.048 &0.012&0.038\\\hline
                   LGU-SLAM(Ours)&0.117&\textbf{0.014}&\textbf{0.022}&\textbf{0.015}&\textbf{0.014} &0.053 &\textbf{0.018} &\textbf{0.020} &\textbf{0.007}&\textbf{0.031}\\              
    \Xhline{.85pt}
			\end{tabular}
		}	
	\end{threeparttable}
\end{table*}

 \subsection{Comparison with state-of-the-art Visual SLAM}
\subsubsection{Results on Visual SLAM challenge benchmark}
The TartanAir Visual SLAM Challenge is a synthetic-data benchmark proposed by the ECCV 2020 SLAM competition, which includes a large number of complex outdoor and indoor scenes, making it extremely challenging for SLAM tasks. We use its official test partitioning to validate LGU-SLAM and compare it with the SOTA algorithm. From Table \ref{table1}, it can be seen that our algorithm performs better than the baseline method DROID-SLAM on the test benchmark, with an average reduction of 0.24 in trajectory errors. The average performance of DPVO is better than ours, thanks to their sparse patch-based correspondence construction in deep visual odometry. Nevertheless, it cannot output dense depth information, making it difficult to achieve high-quality dense mapping for SLAM tasks.

\subsubsection{Results on ETH3D-SLAM benchmark}
ETH3D-SLAM is a testing benchmark for RGB-D SLAM algorithm. Testers need to upload the results to the official for testing and publish the online performance ranking in real time. Our LGU-SLAM is consistent with the baseline method and only trained on tartanAir, without the need for model fine-tuning on the ETH3D-SLAM training set to achieve excellent performance, verifying that our model has good generalization. Fig. \ref{fig6} shows the specific test results, we achieve the second highest score, second only to DVI-SLAM. Compared to the SOTA method GO-SLAM, our AUC improved by 7\% under the "Max\_error=8cm" criterion and 4\% under the "Max\_error=2cm" criterion.

\subsubsection{Results on EuRoC benchmark}
The EuRoC dataset is composed of collected data from real large-scale indoor scenes using drones. We conduct tests on monocular visual odometry and monocular visual SLAM on this dataset, and the results are shown in Table \ref{table2}. It can be seen that our solution performs better on monocular visual odometry than other SOTA methods. Compared to DVI-VO, the average ATE of LGU is decreased by 3\%. Meanwhile, in monocular visual SLAM, our average ATE is less than SOTA algorithms such as DPV-SLAM(by 22\%).

\subsubsection{Results on TUM-RGBD benchmark}
TUM-RGBD is similar to EuRoC, mainly used for small indoor scenes. Referring to previous work, we select the freiburg1 subset as the test set to verify the robustness of our LGU-SLAM. As shown in Table. \ref{table3}, our trajectory prediction error is less than that of the SOTA methods such as GO-SLAM (about by 11\%) and DPV-SLAM (about by 60\%).

\begin{figure}[t!]
	\centering
	\subfloat{
		\includegraphics[width=0.4\textwidth,height=0.25\textwidth]{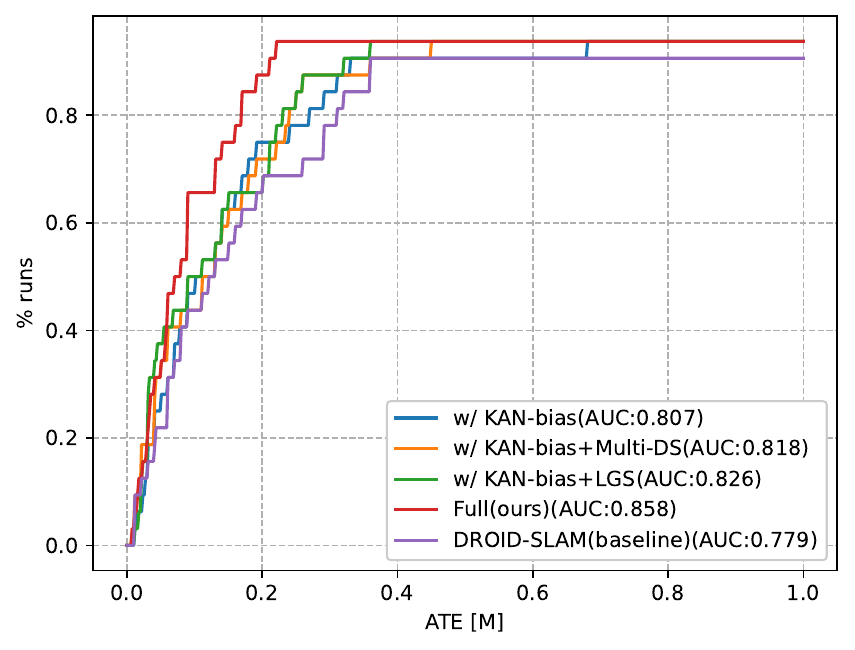}}
	\caption{Ablation study on TartanAir Validation set.(AUC$\uparrow$)}
	\label{fig8} 
\end{figure}

\begin{figure*} [t!]
	\centering
	\subfloat[\label{fig9:a} Comparison of 3D Trajectory in ME001]{
		\includegraphics[width=0.23\textwidth,height=0.19\textwidth]{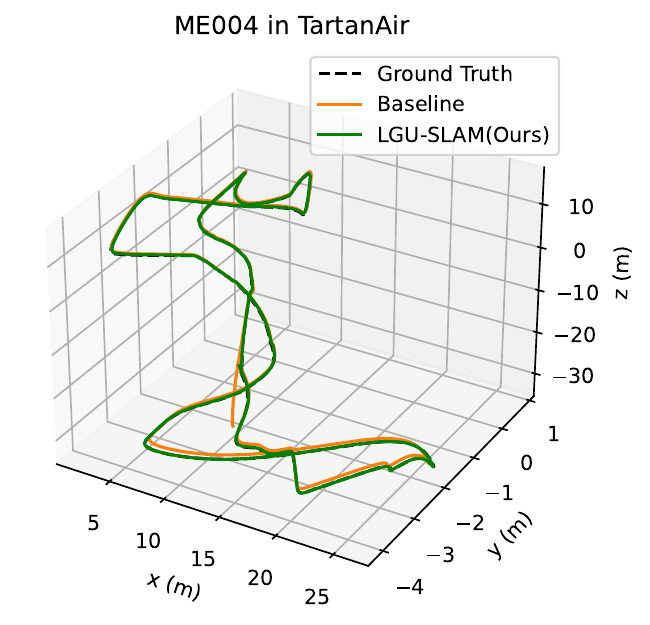}}
	\subfloat[\label{fig9:b} Mapping of ME001 scene]{
			\includegraphics[width=0.48\textwidth,height=0.16\textwidth]{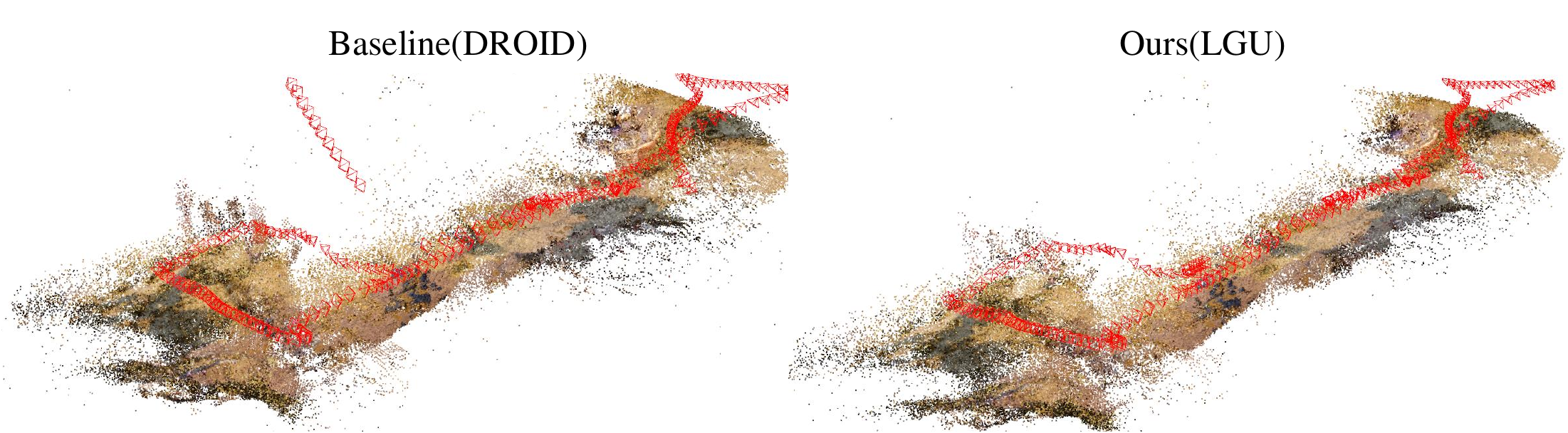}}
       \\
        \subfloat[\label{fig9:d} Comparison of 3D Trajectory in ME004]{
		\includegraphics[width=0.23\textwidth,height=0.19\textwidth]{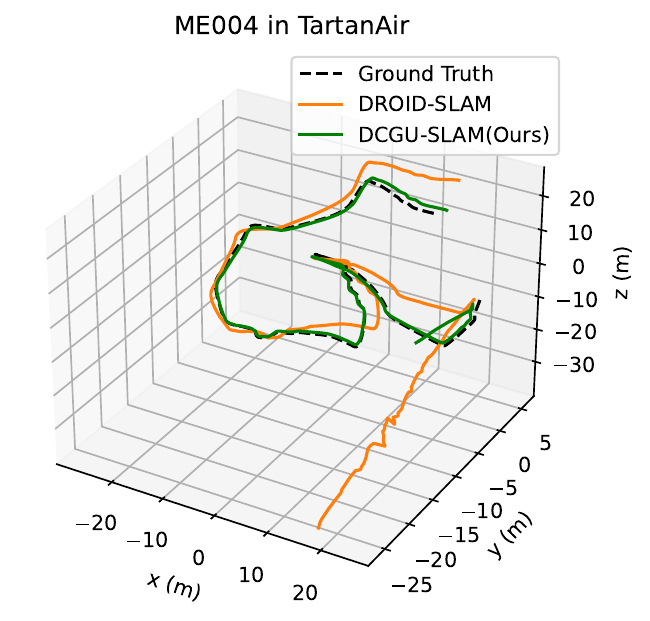}}
	\subfloat[\label{fig9:e}Mapping of ME004 scene]{
			\includegraphics[width=0.48\textwidth,height=0.16\textwidth]{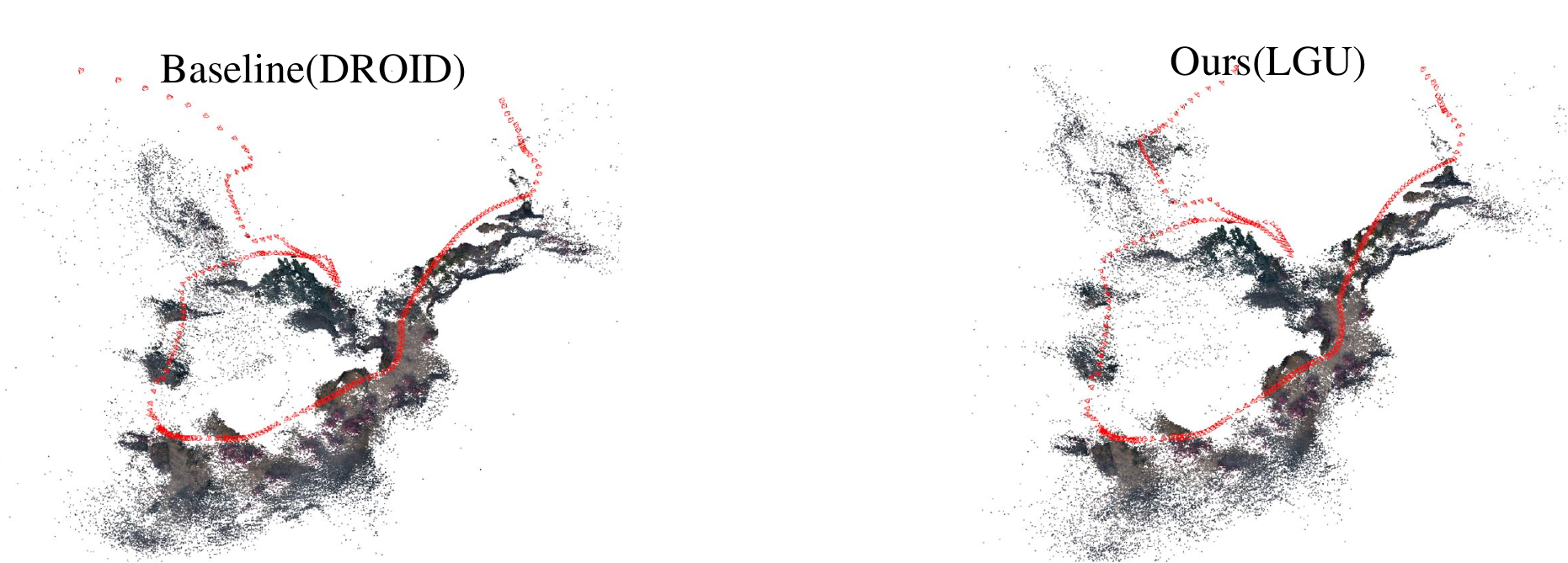}}
	\caption{We selected the challenging scenes "ME001" and "ME004" in the TartanAir to execute a comparative analysis of camera movement trajectory and mapping with the baseline.}
	\label{fig9} 
\end{figure*}

 \subsection{Ablation Study}
The ablation experiment of LGU-SLAM is conducted on the validation set of tartanAir and EuRoC benchmark, and ablation analysis is carried out on the proposed learnable Gaussian uncertainty, multi-scale deformable sampling, and KAN bias GRU.

\subsubsection{Overall Analysis}
\textbf{Effectiveness of proposed methods}. As shown in Fig. \ref{fig8}, we first add KAN bias to the GRU in the temporal iteration model and test it on the TartanAir validation set, namely "w/ KAN-bias". By utilizing the nonlinear learnability of KAN, we achieve complex temporal modeling under limited parameter conditions, and ultimately generate bias terms to regularize the data flow in the temporal iteration. By integrating KAN-bias into GRU, the model's prediction trajectory error is reduced by 16\%(ATE:0.43 to 0.36). Secondly, we test the learnable Gaussian uncertainty with KAN-bias GRU, namely "w/ KAN-bias+ LGS". Compared to the baseline, we further reduce the trajectory error by an average of 25\%(ATE:0.43 to 0.32), which is beneficial for learnable 2D Gaussians. The discrete sampling of the 2D Gaussian is used to weight the area around the distribution center with a radius of $r$. The magnitude of the weight is affected by the covariance of the 2D Gaussian, through end-to-end posterior driving, larger uncertainty and lower attention are appearing in ambiguous areas, and vice versa. Thirdly, we test multi-scale deformable sampling with KAN-bias GRU, namely "w/ KAN-bias+multi-DS". By using spatio-temporal context to generate offsets during the sampling stage ,which enables more efficient utilization of contextual information near sampling points, ultimately achieving lower trajectory error(ATE:0.43 to 0.34). 

Finally, we integrate the full proposed methods and constructed LGU-SLAM. Through orthogonal gain made in both correspondence construction and temporal iteration, our method reduced the average trajectory error by 39\%(ATE:0.43 to 0.26) compared to the baseline. On the tartanAir test set, we also demonstrated stronger robustness than baseline in complex outdoor scenarios, especially in "ME001" and "ME004" scenarios, where LGU-SLAM exhibited more stable tracking performance and lower trajectory errors, as shown in Fig. \ref{fig9}.

\subsubsection{Detail Analysis}
\textbf{Effectiveness of the self-supervised loss}. Firstly, setting a self-supervised loss on the expectation residual not only constrains the optimization of the predicted residual, but also ensures that the center coordinates of each 2D Gaussian are aligned with reliable high visual response points as much as possible, so as to effectively adjust the predicted covariance parameters to describe the uncertainty of the area around the distribution center, thus achieving a closed-loop of the entire learnable Gaussian uncertainty scheme. As shown in Table \ref{table4}, obviously, using self-supervised loss to constrain the optimization of expectation residual and covariance can improve the pose tracking accuracy of Mono VO(ATE:0.46 to 0.34) and SLAM(ATE:0.38 to 0.26) on TartanAir validation set. In the EuRoC benchmark, Our trajectory error has decreased by 2\% and 48\% on VO and SLAM, respectively.

\begin{table}[!ht]
        \caption{Ablation analysis of the soft mask and the self-supervised loss, ATE[m].}
	\centering
	\renewcommand{\arraystretch}{1}
	\setlength{\tabcolsep}{10pt}
	\label{table4}
	\begin{threeparttable}
		\resizebox{\linewidth}{!}{
			\begin{tabular}{l|cc|cc}\Xhline{1.5pt}
				 \multicolumn{1}{c|}{\multirow{3}{*}{Configuration}} & \multicolumn{2}{c|}{\multirow{1}{*}{TartanAir-V}} &\multicolumn{2}{c}{\multirow{1}{*}{EuRoC}}  \\
                \multicolumn{1}{c|}{\multirow{3}{*}{}}&\multicolumn{2}{c|}{\multirow{1}{*}{ATE(m)$\downarrow$}}&\multicolumn{2}{c}{\multirow{1}{*}{ATE(m)$\downarrow$}}\\
                \multicolumn{1}{c|}{\multirow{3}{*}{}}&\multicolumn{1}{c}{\multirow{1}{*}{Mono VO}}&\multicolumn{1}{c|}{\multirow{1}{*}{Mono SLAM}}&\multicolumn{1}{c}{\multirow{1}{*}{Mono VO}}&\multicolumn{1}{c}{\multirow{1}{*}{Mono SLAM}}\\\hline
                baseline&0.57&0.43&0.179&0.024\\
                LGU (w/o SSL)&0.46&0.38&0.142&0.035\\
                LGU (w/o SM)&0.39&0.33&0.146&0.026\\
                LGU (full)&0.34&0.26&0.139&0.018\\
                         
    \Xhline{1.5pt}
			\end{tabular}
		}	
	\end{threeparttable}
\end{table}

\textbf{Effectiveness of the soft mask}. To suppress the ineffective sampling offset in weakly textured areas during offline training of multi-scale deformable sampling, we designed a soft mask based on the visual similarity uncertainty of the original sampling range to filter out redundant information in the generated offset. As shown in Table \ref{table4}, using a soft mask can further reduce the average trajectory error of Mono VO(ATE:0.39 to 0.34) and SLAM(ATE:0.33 to 0.26)  on the TartanAir validation set. In the EuRoC benchmark, Our trajectory error has decreased by 5\% and 30\% on VO and SLAM, respectively.

\begin{table}[!ht]
        \caption{Time and memory usage analysis.}
	\centering
	\renewcommand{\arraystretch}{1}
	\setlength{\tabcolsep}{10pt}
	\label{table5}
	\begin{threeparttable}
		\resizebox{\linewidth}{!}{
			\begin{tabular}{c|c|cc|c}\Xhline{1.5pt}
				 \multicolumn{1}{c|}{\multirow{2}{*}{Dataset}} & \multicolumn{1}{c|}{\multirow{2}{*}{Resolution}} &\multicolumn{2}{c|}{\multirow{1}{*}{Memory}} &\multicolumn{1}{c}{\multirow{2}{*}{Runtime}} \\
                 \multicolumn{1}{c|}{\multirow{2}{*}{}} & \multicolumn{1}{c|}{\multirow{2}{*}{}} & LGU-VO&LGU-SLAM&\multicolumn{1}{c}{\multirow{2}{*}{}}\\\hline
                 TartanAir&$384\times 512$&14G(max) &24G(max)&8(fps)\\
                 EuRoC&$240\times 384$&9G(max) &24G(max)&20(fps)\\
                 ETH3D-SLAM&$312\times 504$&14G(max) &24G(max)&17(fps)\\

    \Xhline{1.5pt}
			\end{tabular}
		}	
	\end{threeparttable}
\end{table}
 \subsubsection{Timing and Memory}
 As shown in Table \ref{table5}, we present the running speed and video memory usage on four datasets: TartanAir, EuRoC, ETH3D, and TUM-RGBD. To optimize the running speed, we adopted Pytorch's CUDA extension to implement corresponding operators for both learnable Gaussian uncertainty and multi-scale deformable sampling, ensuring that the final speed is consistent with DROID-SLAM. Due to the need to construct $H\times W$ Gaussian kernels for correlation volumes in our method, it results in a significant amount of parallel computation, leading to higher memory consumption.

\section{Conclusions}
We propose LGU-SLAM, which mainly consists of learnable Gaussian uncertainty with multi-scale deformable correlation sampling to establish robust correspondence for SLAM. It could ensure dense mapping while enhancing localization accuracy. 
Moreover, we employ dense optical flow fields to correlate two frames, allowing SLAM to predict a dense depth map and providing a viable self-supervised upstream input for current reconstruction methods.
Experiments demonstrate that our LGU-SLAM is effective and superior to compared ones. Its main weaknesses are high memory consumption and low frame rate.  
In the further, investigating a lighter LGU-SLAM scheme with advanced reconstruction has the potential to balance pose tracking and high-quality dense mapping. 






\ifCLASSOPTIONcaptionsoff
  \newpage
\fi

\bibliographystyle{IEEEtran}

 \bibliography{bibtex/bib/IEEEabrv,bibtex/ref}




\end{document}